\documentclass[runningheads]{llncs}
\usepackage{amssymb}
\usepackage{graphicx}
\usepackage{amsmath}
\usepackage{commath}
\usepackage{float}
\usepackage{blindtext}
\usepackage{scrextend}
\begin{document}
\title{On the Applicability of Registration Uncertainty}
%

%
\author{Jie Luo\inst{1,2}
	\and Alireza Sedghi\inst{3}
	\and Karteek Popuri\inst{4}
	\and Dana Cobzas\inst{5} 
	\and Miaomiao Zhang\inst{6} 
	\and Frank Preiswerk\inst{1} 
	\and Matthew Toews\inst{7} 
	\and Alexandra Golby\inst{1} 
	\and Masashi Sugiyama\inst{8,2} 
	\and William M. Wells III\inst{1} 
	\and Sarah Frisken \inst{1}
}  

\authorrunning{J. Luo et al.}

%

\institute{
	Brigham and Women's Hospital, Harvard Medical School, USA 
	\and Graduate School of Frontier Sciences, The University of Tokyo, Japan 
	\and School of Computing, Queen‘s University, Canada 
	\and School of Engineering Science, Simon Fraser University, Canada
	\and Computing Science Department, University of Alberta, Canada 
	\and McKelvey School of Enginerring, Washington University in St.Louis, USA
	\and Ecole de Technologie Superieure, Canada 
	\and Center for Advanced Intelligence Project, RIKEN, Japan\\
	\email{jluo5@bwh.harvard.edu}
}

\maketitle

\begin{abstract}
Estimating the uncertainty in (probabilistic) image registration enables, e.g., surgeons to assess the operative risk based on the trustworthiness of the registered image data. If surgeons receive inaccurately calculated registration uncertainty and misplace unwarranted confidence in the alignment solutions, severe consequences may result. For probabilistic image registration (PIR), the predominant way to quantify the registration uncertainty is using summary statistics of the distribution of transformation parameters. The majority of existing research focuses on trying out different summary statistics as well as means to exploit them. Distinctively, in this paper, we study two rarely examined topics: (1) whether those summary statistics of the transformation distribution most informatively represent the registration uncertainty; (2) Does utilizing the registration uncertainty always be beneficial. We show that there are two types of uncertainties: the transformation uncertainty, $U_\mathrm{t}$, and label uncertainty $U_\mathrm{l}$. The conventional way of using $U_\mathrm{t}$ to quantify $U_\mathrm{l}$ is inappropriate and can be misleading. By a real data experiment, we also share a potentially critical finding that making use of the registration uncertainty may not always be an improvement.

\end{abstract}
\section{Introduction}

Non-rigid image registration is the foundation for many image-guided medical tasks \cite{Maintz,Sotiras}. However, given  the current state of the registration technology and the difficulty of the problem, an uncertainty measure that highlights locations where the algorithm had difficulty finding a proper alignment can be very helpful. Among the approaches that characterize the uncertainty of non-rigid image registration, the most popular, or perhaps the most successful framework is the probabilistic image registration (PIR) \cite{Cobzas,Simpson,Janoos,Risholm,Lotfi,Popuri,Miaomiao,Wassermann,Simpson2,Yang,Heinrich,Folgoc,Miaomiao2,Jax,Adrian}.

In contrast to traditional ``point-estimate" image registration approaches that report a unique set of transformation parameters that best align two images, PIR models transformation parameters as random variables and estimates distributions over them. The mode of the distribution is then chosen as the most likely value of that transformation parameter. PIR has the advantage that the registration uncertainty can be naturally obtained from the distribution of transformation parameters. PIR methods can be broadly categorized into discrete probabilistic registration (DPR) \cite{Cobzas,Lotfi,Popuri,Heinrich} and continuous probabilistic registration (CPR) \cite{Simpson,Janoos,Risholm,Miaomiao,Wassermann,Simpson2,Yang,Folgoc,Miaomiao2,Jax,Adrian,Sedghi}

\subsubsection{Related Work}
Registration uncertainty is a measure of confidence in image alignment solutions. In the PIR literature, the predominant way to quantify the registration uncertainty is using summary statistics of the transformation distribution. Applications of various summary statistics have been proposed in previous research: the Shannon entropy and its variants of the categorical transformation distribution were used to measure the registration uncertainty of DPR \cite{Lotfi}; the variance \cite{Simpson,Yang,Folgoc}, standard deviation \cite{Simpson2}, inter-quartile range \cite{Risholm,Risholm3} and the covariance Frobenius norm \cite{Wassermann} of the transformation distribution were used to quantify the registration uncertainty of CPR. In order to visually assess the registration uncertainty, each of these summary statistics was either mapped to a color scheme, or an object overlaid on the registered image. By inspecting the color of voxels or the geometry of that object, end users can infer the registration uncertainty, which suggests the confidence they can place in the registration result. Utilizing the registration uncertainty is presumably an advantage of PIR \cite{Risholm2,Risholm3,Simpson3}, to date, the majority of existing research focuses on trying out different summary statistics and means to exploit the registration uncertainty.

\subsubsection{Clinical Motivation} In image-guided neurosurgery, surgeons need to correctly understand the registration uncertainty so as to make better informed decisions, e.g., If the surgeon observes a large registration error at location A and small error at location B, without knowledge of registration uncertainty, s/he would most likely assume a large error everywhere and thus entirely ignore the registration. With an accurate knowledge of uncertainty, once the surgeon knows that A lies in an area of high uncertainty while B lies in an area of low uncertainty, s/he would have greater confidence in the registration at B and other locations of low uncertainty. If surgeons are influenced by inaccurate amount of registration uncertainty and place unwarranted confidence in the alignment solutions, severe consequences may result \cite{Risholm,Risholm2,Risholm3}.

The majority of research takes the registration uncertainty for granted. In this paper, we investigate two rarely examined topics: (1) whether summary statistics of the transformation distribution most informatively reflect the registration uncertainty; (2) Does utilizing the registration uncertainty always be beneficial. In Section 2, we identify and discuss two types of uncertainties: the transformation uncertainty $U_\mathrm{t}$ and label uncertainty $U_\mathrm{l}$. By concrete examples, we show that the conventional way of using $U_\mathrm{t}$ to quantify $U_\mathrm{l}$ is inappropriate and can be misleading. In Section 3, by a real data example, we share a potentially critical finding that making use of the registration uncertainty may not always be an improvement. Finally, we summarize in Section 4. It should be noted that registration uncertainty is not equal to registration accuracy. There are excellent works which study standards of registration evaluation \cite{Rohlfing,Fitz,Min}. However, here we focus on the relation among different types of registration uncertainty.


\section{The Ambiguity of Registration Uncertainty}

For illustration purpose, we use DPR in all examples.

\subsection{The DPR Set Up}
In the DPR setting, let $I_\mathrm{t}$ and $I_\mathrm{s}$ respectively be the target and source images $I_\mathrm{t}  ,I_\mathrm{s}: \Omega_I\rightarrow\mathbb{R},\Omega_I\subset\mathbb{R}^d, d=2\: \mathrm{or} \: 3$. The algorithm discretizes the transformation space into a set of $K$ displacement vectors, $\mathcal{D} = \{\mathbf{d}_k\}_{k=1}^K, \mathbf{d}_k\in\mathbb{R}^d$. These displacement vectors radiate from voxels on $I_\mathrm{t}$ and point to their candidate transformation locations on $I_\mathrm{s}$ \cite{Sotiras}. For every voxel $v_i$, the algorithm computes a unity-sum probabilistic vector $\mathcal{P}(v_i)=\{P_k(v_i)\}_{k=1}^K$ as the transformation distribution. $P_k(v_i)$ is the probability of displacement vector $\mathbf{d}_k$. In a standard DPR, the algorithm takes a displacement vector that has the highest probability in $\mathcal{P}(v_i)$ as the most likely transformation $\mathbf{d}_m$. 

Conventionally, the uncertainty of registered $v_i$ is quantified by the Shannon entropy of $\mathcal{P}(v_i)$ \cite{Lotfi}. Since the algorithm takes $\mathbf{d}_m$ as its ``point-estimate", the entropy provides a measure of the extent of dispersion from $\mathbf{d}_m$ of the rest of displacement vectors in $\mathcal{D}$. If other displacement vectors are all as equally likely to occur as $\mathbf{d}_m$, then the entropy is maximal, which indicates that it is completely uncertain which displacement vector should be chosen as the most likely transformation. When the probability of $\mathbf{d}_m$ is much higher than that of other displacement vectors, the entropy decreases, and there is greater certainty that $\mathbf{d}_m$ is the correct choice.

For example, $\mathcal{P}(v_l)=[0.25,0.25,0.25,0.25]$ and $\mathcal{P}(v_r)=[0.1,0.7,0.1,0.1]$ are two discrete transformation distributions. $\mathcal{P}(v_l)$ is uniformly distributed, and its entropy is $E(\mathcal{P}(v_l))=2$. $\mathcal{P}(v_r)$ has an obvious peak, and its entropy is $E(\mathcal{P}(v_r))\approx1.36$, which is lower than $E(\mathcal{P}(v_l))$. For a registered voxel, the entropy of its transformation distribution is usually mapped to a color scheme, clinicians can infer the level of confidence of the registration result by the color of the voxel.

\subsection{Transformation Uncertainty and Label Uncertainty}

In the context of neurosurgery, the goal of image registration is frequently to map the pre-operatively labeled tumor, and/or other tissue, onto the intra-operative patient space for resection. Since registration uncertainty is strongly linked to the goal of registration, here it should also reflect the confidence in the registered labels. However, does the conventional uncertainty measure of DPR, which is the entropy of transformation distribution, truly give insight into the trustworthiness of registered labels? 

\begin{figure}[t]
	\centering
	\includegraphics[height=3.8cm]{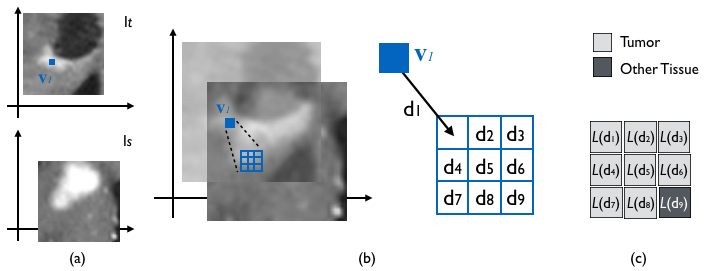}
    \vspace{-2mm}
	\caption{The target image $I_t$ and souce image $I_s$ ; (b) The discretized transformation space $\mathcal{D}$; (c) The corresponding tissue label $L(\mathbf{d}_k)$ for $\mathcal{D}$. .} 
	\vspace{-4mm}
	\label{fig1}
\end{figure}

In a hypothetical DIR example, $I_t$ and $I_s$ in Fig.1(a) are the intra-operative target and pre-operative source images, respectively. Voxel $v_1$ on $I_t$ is the voxel we want to register. In Fig.1(b), we can see that the discretized transformation space $\mathcal{D} = \{\mathbf{d}_k\}_{k=1}^9$ is a set of nine displacement vectors. Each displacement vector is linked to a candidate corresponding voxel of $v_1$. The labels $L(\mathbf{d}_k)$ are for voxels associated with $\mathbf{d}_k$. In this example, there are labels for the tumor and other tissue, as shown in Fig.1(c).

Fig.2 shows a transformation distribution $\mathcal{P}(v_1)=\{P_k(v_1)\}_{k=1}^9$  and its bar chart. We observe that $P_5(v_1)$ has the highest probability in $\mathcal{P}(v_1)$; therefore, $\mathbf{d}_5$'s corresponding label, $L(\mathbf{d_5})=\mathrm{Tumor}$, will be assigned to the registered $v_1$.

Although $\mathcal{P}(v_1)$ has its mode at $P_5(v_1)$, the entire distribution is more or less uniformly distributed.  The entropy of $\mathcal{P}(v_1)$, $E(\mathcal{P}(v_1))\approx3.15$, is close to the maximum. Therefore, the conventional uncertainty measure will suggest that the registration uncertainty of $v_1$ is very high and highlight it with a bright color. Upon noticing the high degree of uncertainty in registered $v_1$, surgeons would place less confidence in its tumor label and make surgical plans accordingly.

\vspace{-3mm}
\begin{figure}[H]
	\centering
	\includegraphics[height=2.5cm]{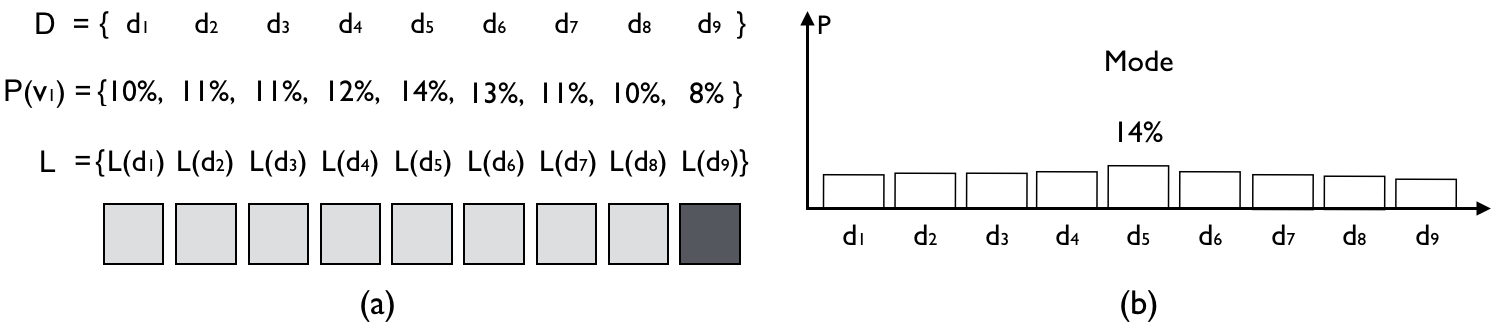}
	\vspace{-3mm}
	\caption{(a) $\mathcal{P}(v_1)$ and corresponding labels; (b) The bar chart of $\mathcal{P}(v_1)$.}
	\vspace{-4mm}
	\label{fig2}
\end{figure}

On the other hand, let us take into account the label $L(\mathbf{d}_k)$ associated with each $\mathbf{d}_k$ and form a label distribution. As shown in Fig.3(a), even if $\mathbf{d}_1,\dots,\mathbf{d}_8$ are different displacement vectors, they correspond to the same label as the most likely displacement vector $\mathbf{d}_5$.  If we accumulate the probability for all labels in $\mathcal{L}$, it is clear that ``tumor" is the dominant one. Interestingly, despite being suggestive of having high registration uncertainty using the conventional uncertainty measure, the label distribution in Fig.3(b) indicates that it is quite trustworthy to assign a tumor label to the registered $v_1$. In addition, the entropy of the label distribution is as low as 0.4, which also differs from the high entropy value computed from the transformation distribution.

\begin{figure}[t]
	\centering
	\includegraphics[height=3.2cm]{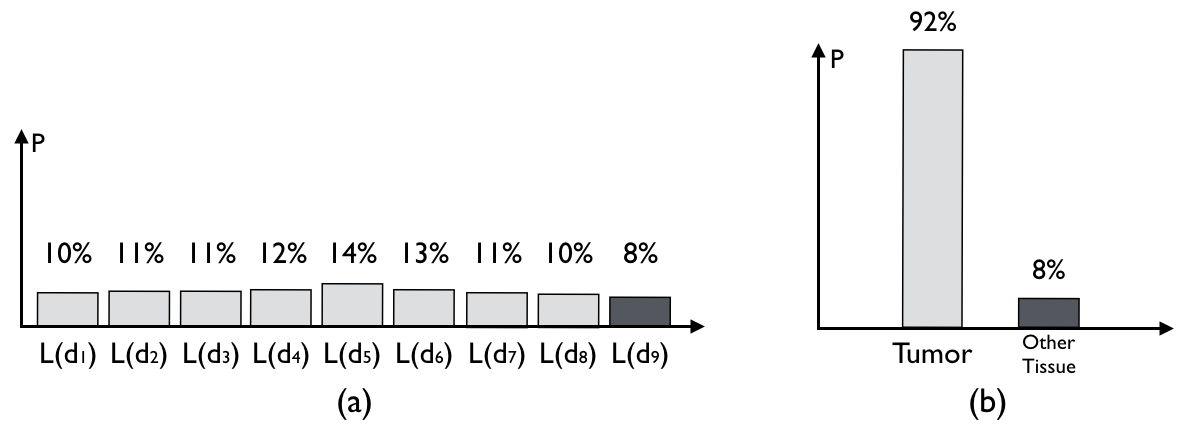}
	\vspace{-3mm}
	\caption{(a) Bar chart of the transformation distribution $\mathcal{P}(v_1)$ taking into account $L(\mathbf{d_k})$; (b) The label distribution of the registered $v_1$.}
	\label{fig:correct2}
	\vspace{-4mm}
\end{figure}

In the example above, there appear to be two kinds of uncertainty. We name the uncertainty computed from the transformation distribution as the transformation uncertainty $U_\mathrm{t}$, and the uncertainty relating to the goal of registration as label uncertainty $U_\mathrm{l}$. Examples of $U_\mathrm{l}$ can be uncertainty in a categorical classification, or uncertainty in the intensity value of registered voxels.

In the PIR literature, the definition of registration uncertainty is ambiguous, because researchers do not differentiate $U_\mathrm{t}$ from $U_\mathrm{l}$, and perhaps subconsciously use $U_\mathrm{t}$ to quantify $U_\mathrm{l}$. The previous counter-intuitive example demonstrates that high $U_\mathrm{t}$ does not guarantee high $U_\mathrm{l}$. In fact, the value of $U_\mathrm{t}$ can barely guarantee any useful information at all about the $U_\mathrm{l}$.

More precisely, for point-estimate image registration, let $\Omega_T$ be the set of all estimated transformation, and $\Omega_L$ be the set of all possible corresponding labels (categorical labels or intensity values). The algorithm assigns a transformation $t\in \Omega_T$ to a voxel. By a non-linear function $f_{point}: \Omega_T\rightarrow \Omega_L$, the voxel will have its label $l\in \Omega_L$ as: 
\begin{equation}
l=f_{point}(t).
\end{equation}
In this case, the function $f_{point}$ is surjective, and $t$ always has a unique corresponding $l$.  However, in the PIR setting, the voxel transformation becomes a random variable $T$. The corresponding label $L$ is a function of $T$: 
\begin{equation}
L=f_{prob}(T).
\end{equation}
therefore, it is also a random variable. Even if $T$ and $L$ are intuitively correlated, given different image context, there is no guaranteed analytical way to compute the uncertainty propagation from $T$ to $L$. Thus it's inappropriate to measure the uncertainty of $L$, by the summary statistics of $T$.

In the registration community, researchers routinely distinguish between intensity match and e.g. DICE scores. It also makes sense to distinguish the transformation uncertainty and label uncertainty. In practice, $U_\mathrm{t}$ and $U_\mathrm{l}$ around certain areas, i.e., the tumor boundary, is quite dissimilar. Propagating $U_\mathrm{t}$ to the surgeon, as if it is the $U_\mathrm{l}$ can mislead them to place unwarranted confidence in the alignment solution and result in severe consequences.

\begin{figure}[t]
	\centering
	\includegraphics[height=5.3cm]{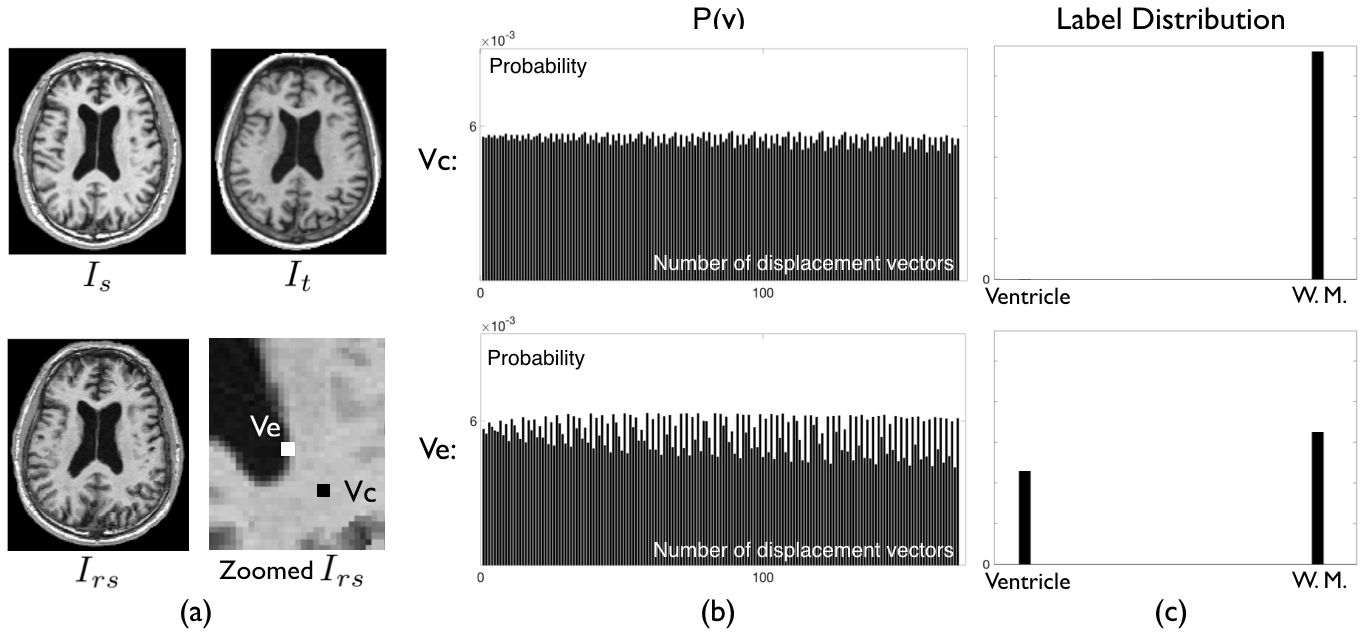}
	\vspace{-2mm}
	\caption{(a) Input and result of the CUMC12 data example,  $v_c$ and $v_e$ are two voxels of interest on the registered source image; (b) The transformation distribution of $v_c$ and $v_e$ in the DIR; (c) Label distributions of registered $v_c$ and $v_e$.}
	\vspace{-6mm}
	\label{fig:correct2}
\end{figure}

\subsubsection{Real data examples} 

As shown in Fig.4, $I_\mathrm{t}$ and $I_\mathrm{s}$ are two brain MRI images arbitrarily chosen from the CUMC12 dataset. Subsequent to performing a DIR, we obtained the registered source image $I_\mathrm{rs}$. The goal of this registration was to determine the categorical label, whether it is a ventricle or a white matter, for registered voxels of interest $v_c$ and $v_e$. The transformation distribution of $v_c$ is more uniformly distributed than that of $v_e$. Therefore, conventional entropy-based methods will report $v_c$ as having higher registration uncertainty than $v_e$. However, as we form a label distribution in Fig.4(c),  it is clear that $v_e$, despite having a lower $U_\mathrm{t}$, is assigned a label that is more uncertain. Examples that demonstrate the dissimilarity between $U_\mathrm{t}$ and $U_\mathrm{l}$ can be frequently found in the registration of various kind of images.

\vspace{-2mm}
\section{Credibility of Label Distribution}

Utilizing the registration uncertainty, in particular the full label distribution, to benefit registration-based tasks is presumably an advantage of PIR. Many research of registration uncertainty reported positively over its impact in applications \cite{Risholm2,Risholm3,Simpson3}. However, to the best of our knowledge, there does not exist any validation study about whether we should use the registration uncertainty. In this section, we design an experiment to explore whether utilizing the registration uncertainty always results in an improvement. 

 In PIR, the registered voxel has the corresponding label of the most likely transformation $L(\mathbf{d}_m)$, upon which the registration evaluation is also based \cite{Cobzas,Simpson,Janoos,Risholm,Lotfi,Popuri,Miaomiao,Wassermann,Simpson2,Yang,Heinrich,Folgoc,Miaomiao2,Jax}. Likewise, we can derive the most likely label $L_m$ from the full label distribution. If utilizing the registration uncertainty, like reported in previous research, always be beneficial, then $L_m$ should be always better than $L(\mathbf{d}_m)$.

In the following pilot experiment: an MRI image is arbitrarily chosen from the BRATS dataset \cite{BRATS} and synthetically deformed.Then we registered the original data with the deformed data using DIR. By doing so, we know the ground truth intensity for every registered voxel so that we can compare whether it is $L(\mathbf{d}_m)$ or  $L_m$ closer to the ground truth.

\begin{figure}[t]
	\centering
	\includegraphics[height=5.8cm]{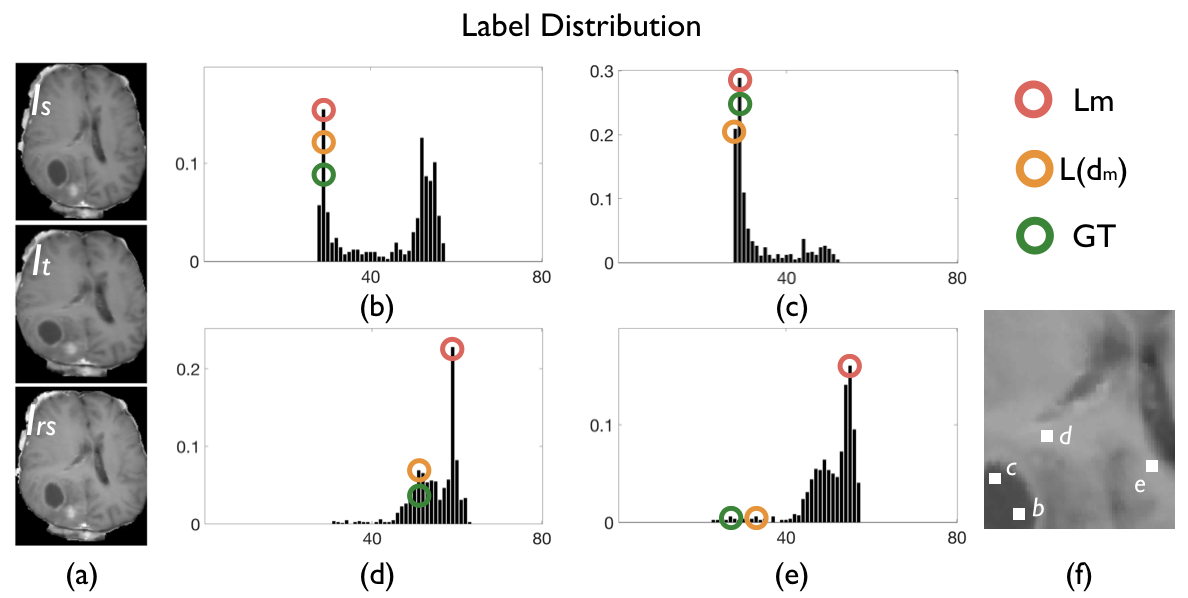}
	\vspace{-3mm}
	\caption{(a) Input and result of the registration example; (b,c,d,e) Intensity label distributions of voxels $v_b$,$v_c$,$v_d$ and $v_e$; (f) Approximate locations of tested voxels.}
	\vspace{-5mm}
	\label{fig:useful}
\end{figure}

Here we are interested in the intensity label distributions of four registered voxels $v_\mathrm{b},v_\mathrm{c},v_\mathrm{d}$ and $v_\mathrm{e}$, shown in Fig.5(b), (c), (d), and (e) respectively. In Fig.5, the red circle indicates the most likely intensity label $L_m$ given by the full transformation distribution, the orange circle indicates the corresponding intensity label of the transformation mode $L(\mathbf{d}_m)$, and the green circle is the Ground Truth (GT). We observe that for $v_b$, $L_m$ and $L(\mathbf{d}_m)$ are both equal to the GT. On the other hand, $L_m$ and  $I(\mathbf{d_m})$ for $v_\mathrm{c}, v_\mathrm{d}$ and $v_\mathrm{e}$, are not the same. As seen in Fig.5(c), the $L_m$ of the registered $v_c$ is equal to the GT intensity, and is more accurate than $I(\mathbf{d_m})$. Yet, unexpectedly, for $v_\mathrm{d}$ and $v_\mathrm{e}$, their $I(\mathbf{d}_m)$ is closer to the GT than their $L_m$. Voxels such as $v_d$ and $v_e$ were found frequently in our experiments using other real data. This surprising result indicates that utilizing the full transformation distribution can actually give a poorer/less accurate estimation than using the transformation mode alone.

Researchers have attempted to present the visualized full label distribution of functional areas in fMRI to neurosurgeons \cite{Risholm3}. However, based on the above finding, if $L_m$ can give poorer estimation, the full label distribution might also have questionable credibility. Conveying such false information to surgeons would certainly be detrimental to the outcome of surgery.

It is noteworthy that in PIR, the estimation of $T$ and $L$ is influenced by the choice of hyper parameters, priors, and image context. Other PIR approaches can yield different findings. Nevertheless, studying the credibility of the label distribution before using it in practice warrants increased investigation.

\section{Discussion}

The majority of research takes the registration uncertainty for granted. We summarize current approaches of quantifying registration uncertainty and point out some fundamental problems which would make researchers rethink, or even re-work approaches for quantifying and applying registration uncertainty. 

At this stage, even the uncertainty is a useful addition to the registration result, we recommend treating it with caution: (1) It is advised to distinguish $U_\mathrm{t}$ and $U_\mathrm{l}$ in applications. Instead of using the unified term "registration uncertainty", i.e., we can use $U_\mathrm{t}$ to indicate the confidence for a predicted instrument location in neurosurgery; (2) Since the credibility of label distribution is unclear, we should avoid using $U_\mathrm{l}$ in clinical settings and put further effort in studying the implication of PIR results. We believe that this paper will serve as a foundation and draw more attention to this topic.

\subsubsection{Acknowledgement}

MS was supported by the International Research Center for
Neurointelligence (WPI-IRCN) at The University of Tokyo Institutes for
Advanced Study. This work was also supported by NIH grants P41EB015898, P41EB015902 and 5R01NS049251.

%
%
%

\end{document}